\documentclass[cameraready]{Interspeech}

\title{Language-Aware Distillation for Multilingual Instruction-Following Speech LLMs with ASR-Only Supervision}

\author[affiliation={1}, orcid=0009-0000-8699-3362]{Shreyas}{Gopal}
\author[affiliation={1, 2}, orcid=0009-0004-3597-8284]{Donghang}{Wu}
\author[affiliation={1}, orcid=0000-0001-8883-7494]{Ashutosh}{Anshul}
\author[affiliation={1, 3}, orcid=0009-0001-0884-8888]{Yeo}{Yue Heng}
\author[affiliation={1}, orcid=0000-0002-5718-570X]{Yizhou}{Peng}
\author[affiliation={1}]{Haoyang}{Li}
\author[affiliation={1}, orcid=0000-0002-3998-9229, correspondingauthor]{Hexin}{Liu}
\author[affiliation={1}, orcid=0000-0001-6257-7399]{Eng Siong}{Chng}

\address{
    $^1$ College of Computing and Data Science, Nanyang Technological University, Singapore \\
    $^2$ AI Singapore, National University of Singapore \\
    $^3$ Institute for Infocomm Research (I$^2$R), A$^{*}$STAR, Singapore
}

\email{shreyas011@ntu.edu.sg; hexin.liu@ntu.edu.sg}


\usepackage{comment}
\usepackage{multirow}
\usepackage{array}
\newcolumntype{C}[1]{>{\centering\arraybackslash}p{#1}}
\newcolumntype{L}[1]{>{\raggedright\arraybackslash}p{#1}}

\begin{document}

\maketitle

\vspace{-4pt}
\begin{abstract}
    Speech Large Language Models (LLMs) that understand and follow instructions in many languages are useful for real-world interaction, but are difficult to train with supervised fine-tuning, requiring large, task-specific speech corpora. While recent distillation-based approaches train performant English-only Speech LLMs using only annotated ASR data by aligning text and speech using only a lightweight projector, these models under-perform when scaled to multilingual settings due to language interference in the shared projector. We address this by introducing language-aware distillation using a query bank and a gating network that selects or mixes query tokens using a Q-Former projector. Our approach shows gains of 14\% over matched multilingual distillation baselines on instruction following. We further synthesize Audio--MLQA, a multilingual spoken QA benchmark built on MLQA with high-quality TTS questions. Our best model improves over existing Speech LLM baselines by 32\% on Audio-MLQA.
\end{abstract}





\section{Introduction}

Speech Large Language Models (Speech LLMs) integrate a speech encoder with a text LLM to support speech-conditioned understanding, reasoning, and generation. Cascaded ASR-to-LLM pipelines convert speech signals to text early in the ASR stage, leading to information beyond semantics being discarded. In contrast, existing end-to-end approaches~\cite{wang2024blsp_emo, wang2025benchmarking, nguyen2023generative} show that Speech LLMs can preserve paralinguistic information and capture emotional cues. However, many recent Speech LLMs rely on multi-stage training that combines large-scale pretraining with supervised fine-tuning (SFT)~\cite{Qwen-Audio,chu2024qwen2,tang2024salmonn,chen2024salm}. Other efforts scale multilingual coverage through large training distributions and specialized regional models~\cite{he2025meralion,SeaLLMs-Audio,fang-etal-2024-llama-omni}. While effective, these pipelines require substantial training resources for SFT and commonly fine-tune the speech encoder or the LLM, which can lead to poor generation due to catastrophic forgetting~\cite{hsiao2025forgetting,held-etal-2025-distilling} while biasing toward recent training data. The problem is exacerbated in multilingual settings, where annotated data is inherently low-resource, annotation is uneven across languages, and task-specific data is almost non-existent.

In such conditions, a common approach is to generate synthetic task-specific training data using text-to-speech (TTS) systems~\cite{majumder2024tango,noroozi2024instruction}, which themselves require large amounts of data to train. Alternatively,~\cite{dao2025speechless} propose alignment of discrete speech and text tokens to generate pseudo-labels from text for non-English languages. However, their supervision quality is limited by the reliability and coverage of the pseudo-label generator. Alternatively, to reduce training costs and avoid reliance on task-specific data, some recent studies focus on speech--text alignment rather than large-scale SFT. DiVA~\cite{held-etal-2025-distilling} adapts a frozen LLM to speech using paired ASR data via context distillation. Similarly, BLSP~\cite{wang2023blsp} encourages consistent generation when conditioned on speech versus the transcription, improving semantic alignment via continuation writing.

Extending alignment-based Speech LLMs to multilingual settings reveals critical architectural limitations in current projection methods. Prior speech-to-LLM works~\cite{held-etal-2025-distilling,dong2024_interspeech,shang2024_interspeech} use a trainable Query-Transformer (Q-Former)~\cite{li2023blip} style projector that learns a static sequence of query tokens to align speech embeddings with transcript text embeddings. While this shared projector facilitates some transfer across related languages, we observe that a single, static query sequence proves insufficient for capturing the distinct phonetic and semantic nuances as the number and diversity of languages increases. Particularly for distant pairs such as English and Chinese, this lack of language-specific flexibility leads to performance degradation due to language interference, where dominant languages in the training distribution can overshadow lower-represented ones in the shared representation space. This aligns with multilingual ASR findings where explicit language conditioning is often beneficial~\cite{whisper,aaron_2stage}, and language-aware routing and gating have also shown reduced interference in multilingual systems~\cite{wang2023language,li-etal-2023-adaptive-gating}.

In this study we demonstrate how to extend ASR-only distillation and behavior alignment toward robust multilingual speech understanding by introducing a language-aware framework. Our aim is to train performant multilingual Speech LLMs efficiently (using only 5.8K hours of data in total to support 6 languages) and with minimal additional trainable capacity, keeping the speech encoder and the LLM frozen. We also share high-quality multi-lingual open-ended instruction following and close-ended spoken question answering evaluation datasets to benefit future research in this direction.

The main contributions of this work are:
\begin{itemize}
\item We propose a novel language-aware distillation method for multilingual Speech LLMs requiring significantly fewer ASR-only training resources
\item We show consistent gains over a matched multilingual baseline and external models on open-ended instruction following and close-ended spoken QA tasks
\item We provide multilingual open-ended and close-ended evaluation data to support future benchmarking
\end{itemize}

\begin{figure*}[t]
  \centering
  \includegraphics[width=0.96\textwidth]{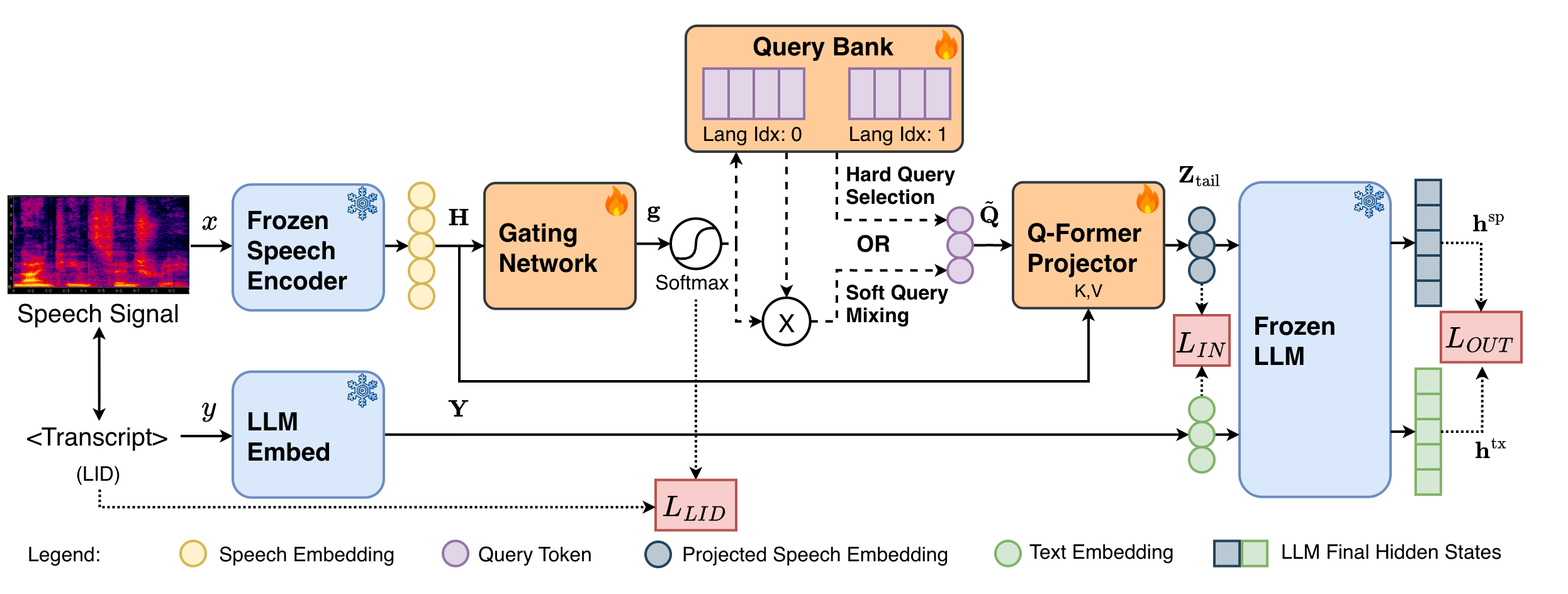}
  \vspace{-10pt}
  \caption{End-to-end model architecture. Trainable components are colored orange.}
  \label{fig:model_architecture}
  \vspace{-12pt}
\end{figure*}

\section{Methodology}

Figure~\ref{fig:model_architecture} illustrates our approach which consists of 4 components: (1) a frozen speech encoder, (2) a Q-Former projector, (3) a frozen LLM and (4) the query-selection module.

\subsection{Model Architecture}

Building on prior Speech LLM models, we train a lightweight adapter using paired speech and transcript data. Given speech $x$ and transcript $y$, the model learns a cross-modal projector that converts speech embeddings into \emph{text-like} representations that act as a \emph{soft speech prefix} for the frozen LLM to generate text.

\vspace{4pt}
\textbf{Frozen Speech Encoder}: We use the frozen Whisper-large-v3 encoder~\cite{whisper} to compute speech embeddings $\mathbf{H}=E_{\text{sp}}(x)\in\mathbb{R}^{T\times d_s}$, where $d_s$ is the hidden size and  $T$ is the number of time steps, and $x$ stands for the speech input.

\vspace{4pt}
\textbf{Frozen LLM and Text Embeddings}: We use a frozen Llama-SEA-LION-v3-8B-IT~\cite{ng2025sea} as the text backbone. It is comparable in scale to Llama3-8B~\cite{grattafiori2024llama} and provides better coverage for several Southeast Asian languages especially low-resource languages (see Table~\ref{tab:main_results_allinone}). Given tokens of text transcriptions, $\hat y=(\hat{y}_1,\dots,\hat{y}_N)$, we obtain their input embeddings from the frozen LLM embedding layer, $\mathbf{Y}=\text{Emb}(\hat{y})\in\mathbb{R}^{N\times d_\ell}$. We use $\mathbf{Y}$ as the teacher signal for input distillation and for defining the text-conditioned teacher signal used in output distillation (\S\ref{loss-fns}). Keeping the LLM frozen effectively prevents catastrophic forgetting of reasoning ability.

\vspace{4pt}
\textbf{Modality Adapter}: The projector attends over $\mathbf{H}$ and produces $L$ projected embeddings
$\mathbf{Z}=P_{\theta}(\mathbf{Q},\mathbf{K},\mathbf{V})\in\mathbb{R}^{L\times d_\ell}$, with $(\mathbf{K},\mathbf{V})=f(\mathbf{H})$,
where $\mathbf{Q}\in\mathbb{R}^{L\times d_\ell}$ are learnable query tokens. The output $\mathbf{Z}$ acts as a soft speech prefix for the frozen LLM. We initialize queries with $\mathbf{Q}\sim\mathcal{N}(0,0.02^2)$, and projector weights from whisper-large-v3 decoder following~\cite{held-etal-2025-distilling}.

\vspace{-2pt}
\subsection{Language-Aware Distillation}
Though the shared Q-Former projector shows some extent of cross-lingual transfer for related languages, a single shared query sequence can struggle as the number and diversity of languages increases. In practice, languages that dominate the data distribution (and their nearby languages) tend to improve, while distant and lower-represented languages degrade due to interference. Hence, we introduce a \emph{query bank} and a \emph{gating network} that selects or mixes queries conditioned on the input speech to better disentangle language-specific information.

\vspace{-2pt}
\subsubsection{Query Bank and Gating Network}
Let $K$ be the number of languages. We maintain a bank of learnable query tokens $\mathcal{B}=\{\mathbf{Q}^{(k)} \in \mathbb{R}^{L\times d_{\ell}}\}_{k=1}^{K}$, where $\mathbf{Q}^{(k)}$ denotes the query sequence for language $k$. The projector is conditioned on effective query input $\tilde{\mathbf{Q}}$, obtained by mixing queries across languages or selecting a single language query.

Given speech embeddings $\mathbf{H}\in\mathbb{R}^{T\times d_s}$, the gating network outputs language logits $\mathbf{g}=G_{\phi}(\mathbf{H})\in\mathbb{R}^{K}$. We implement $G_{\phi}$ using either (i) a convolutional LID-style network with temporal down-sampling followed by pooling, or (ii) a lightweight attention-pooling MLP that pools $\mathbf{H}$ into an utterance vector. Both variants produce the same logits $\mathbf{g}$ and differ only in how the utterance-level summary is computed. The logits $\mathbf{g}$ are then used for soft query mixing or hard query selection to prepare $\tilde{\mathbf{Q}}$. The performance of both query selection methods (soft vs. hard), and gating designs (conv vs. attn) are compared in~\S~\ref{sec:ablation_studies}.

\vspace{4pt}
\textbf{Soft Query Mixing}:
The mixture weights are computed as $\boldsymbol{\pi}=\mathrm{softmax}(\mathbf{g})$ and the mixed query is formed as $\tilde{\mathbf{Q}}_{\text{soft}}=\sum_{k=1}^{K}\pi_k\,\mathbf{Q}^{(k)}$. This allows smooth information sharing across related languages while retaining language-specific features and emphasis through $\boldsymbol{\pi}$.

\vspace{4pt}
\textbf{Hard Query Selection}:
We select a single language index $k^*=\arg\max_k g_k$ and use $\tilde{\mathbf{Q}}_{\text{hard}}=\mathbf{Q}^{(k^*)}$ in the forward pass. To stabilize training, we use a straight-through estimator and backpropagate through the soft mixture:
$\tilde{\mathbf{Q}} \leftarrow \tilde{\mathbf{Q}}_{\text{hard}} + \big(\tilde{\mathbf{Q}}_{\text{soft}}-\mathrm{stopgrad}(\tilde{\mathbf{Q}}_{\text{soft}})\big)$.
This keeps selection discrete for inference, providing significant emphasis for each language, but allows the gate to learn from downstream distillation losses.

\vspace{-2pt}
\subsubsection{Scheduled Teacher Forcing}
Each training sample is loosely annotated with a language label $\ell$. Early on, due to random query initialization, we apply scheduled teacher forcing for query selection to stabilise training. At step $s$, with probability $p_{\text{tf}}(s)$ we set $k^* \leftarrow \ell$; otherwise we use the model prediction. We anneal $p_{\text{tf}}(s)$ from $1$ to $0$ with a cosine schedule such that $p_{\text{tf}}=0$ at $50\%$ of the training steps.

\subsection{Training Objective}

\label{loss-fns}
The loss function consists of three components: language identification loss, input distillation loss, and output distillation loss.

\vspace{4pt}
\textbf{Language identification loss}: We supervise the gating network using cross-entropy on the language logits $\mathbf{g}\in\mathbb{R}^{K}$ with language labels $\ell$. We ignore samples with unknown labels (encoded as $-1$). Let $\mathcal{I}=\{b \mid \ell_b \ge 0\}$ denote the set of valid samples in a batch with size $B$. The LID loss is
\begin{equation}
\mathcal{L}_{\text{LID}} =
-\frac{1}{|\mathcal{I}|}\sum_{b\in\mathcal{I}}
\log \mathrm{softmax}(\mathbf{g}_b)_{\ell_b}
\end{equation}

\textbf{Input distillation loss}: We encourage the projected speech embeddings $\mathbf{Z}$ to match the transcript-derived LLM input embeddings $\mathbf{Y}$. Since the speech sequence is typically longer than the transcript, we align the first $T$ transcript embeddings with the last $T$ projected speech embeddings (the ``audio tail''), i.e., $\mathbf{Y}_{\text{head}}=\mathbf{Y}_{1:T}$ and $\mathbf{Z}_{\text{tail}}=\mathbf{Z}_{L-T+1:L}$. The loss is only computed over valid (non-padding) transcript positions using a mask $\mathbf{m}\in\{0,1\}^{T}$, where $N_b=\sum_{t=1}^{T} m_{b,t}$. Using $L_2(\cdot,\cdot)$ to denote $L_2$ distance, the input distillation loss is:
\begin{equation}
\mathcal{L}_{\text{IN}} =
\frac{1}{B}\sum_{b=1}^{B}
\frac{1}{\max(1, N_b)}
\sum_{t=1}^{T} m_{b,t}\,
L_2\left(\mathbf{Z}^{(b)}_{\text{tail},t}, \mathbf{Y}^{(b)}_{\text{head},t}\right)
\end{equation}

\textbf{Output distillation loss}: In order to match LLM behavior under speech versus transcript conditioning, we align the final hidden representations produced by the frozen LLM. Let $\mathbf{h}^{\text{sp}}_{b,t}\in\mathbb{R}^{d_\ell}$ be the last-layer hidden state at position $t$ when the LLM is conditioned on speech (via $\mathbf{Z}$), and let $\mathbf{h}^{\text{tx}}_{b}\in\mathbb{R}^{d_\ell}$ be the corresponding teacher vector obtained from transcript-only conditioning (detached from gradients). For each sample we extract the hidden state at the last non-padding token index $t_b$. The output distillation loss is
\begin{equation}
\mathcal{L}_{\text{OUT}} =
\frac{1}{B}\sum_{b=1}^{B}
L_2\left(\mathbf{h}^{\text{sp}}_{b,t_b},\ \mathbf{h}^{\text{tx}}_{b}\right)
\end{equation}
$L_{IN}$ provides coarse prefix--text embedding alignment, avoiding frame alignment~\cite{held-etal-2025-distilling}. $L_{OUT}$ matches speech/transcript-conditioned LLM hidden states. The final loss function is:
\begin{equation}
    \mathcal{L} = \lambda_{\text{IN}}\mathcal{L}_{\text{IN}} + \lambda_{\text{OUT}}\mathcal{L}_{\text{OUT}} + \lambda_{\text{LID}}\mathcal{L}_{\text{LID}},\\
\end{equation}
where $\lambda_{\text{IN}}$, $\lambda_{\text{OUT}}$ and $\lambda_{\text{LID}}$ are the loss weights.

\section{Experiments}

\vspace{-2pt}
\subsection{Training Data}

Only annotated ASR data was used during the training of the model. The various sources and amounts of data are detailed in Table~\ref{tab:train_data_breakdown_dataset}. We filter out audio samples with word error rate (WER\%) greater than 10\% from expert ASR models. 

\vspace{-2pt}
\subsection{Evaluation Data}

\begin{table}[t]
  \caption{Breakdown of ASR training data by language, number of hours, and sample count. CV is CommonVoice.}
  \label{tab:train_data_breakdown_dataset}
  \centering
  \footnotesize
  \setlength{\tabcolsep}{3pt}
  \renewcommand{\arraystretch}{0.95}
  \begin{tabular}{l l r r}
    \toprule
    \textbf{Dataset} & \textbf{Language} & \textbf{Hours} & \textbf{\#Samples} \\
    \midrule
    CV21~\cite{commonvoice:2020} & EN & 1750 & 1.10M \\
    ViVoice~\cite{capleaf_vivoice_2024} & VI & 900 & 0.785M \\
    YODAS2~\cite{li2023yodas} & ID & 950 & 0.996M \\
    MagicData~\cite{magicdata_2019} & ZH & 755 & 0.500M \\
    CV24~\cite{commonvoice:2020} & ZH & 100 & 0.030M \\
    CV24~\cite{commonvoice:2020} & ES & 515 & 0.357M \\
    CV24~\cite{commonvoice:2020} & DE & 900 & 0.569M \\
    \midrule
    \textbf{Total} & & \textbf{5870} & \textbf{4.33M} \\
    \bottomrule
  \end{tabular}
  \vspace{-14pt}
\end{table}

Due to the lack of established evaluation benchmarks for multilingual speech understanding tasks, we have also developed in-house synthetic evaluation data to support our findings in addition to existing benchmarks. Data used for evaluation is ``unseen" during training with no speaker overlap in training and evaluation data. All high-quality synthetic audio is generated at 44.1KHz using state-of-the-art commercially available TTS.

\vspace{-2pt}
\subsubsection{Open-Ended Evaluation Dataset}
We evaluate three languages in the open-ended benchmark and test the ability of the model to understand the spoken instruction and generate a relevant response. For Chinese (ZH), we use the \emph{AlpacaEval-zh} split from UROBench~\cite{yan2025uro}, which contains 147 open-ended instruction-following samples derived and translated from VoiceBench~\cite{chen2024voicebench}. For English (EN), we use the ALPACA-Audio and OpenHermes-Audio subsets from AudioBench~\cite{wang-etal-2025-audiobench}, each containing 100 samples (200 total). For Indonesian (ID), we translate the English ALPACA-Audio and OpenHermes-Audio prompts into Indonesian with the help of expert native speakers, and synthesized the corresponding audio prompts using state-of-the-art TTS systems with a fixed pool of 2 male and 2 female high-quality voices.

\vspace{-2pt}
\subsubsection{Close-Ended Evaluation Dataset - Audio--MLQA}
To evaluate whether the model can align spoken questions with key words and phrases from textual context, we build a close-ended spoken QA benchmark; Audio-MLQA. The benchmark can be used to evaluate 5 languages: English (EN), Vietnamese (VI), Spanish (ES), German (DE), and Chinese (ZH). We source text context and questions from MLQA~\cite{lewis2019mlqa}, which provides multilingual textual QA data. We randomly sample 250 items per language and synthesize the spoken questions using state-of-the-art TTS systems. To reduce the speaker-specific bias, we use a fixed pool of 10 voices per language (5 male and 5 female) and randomly assign one voice to each sample.

\vspace{-2pt}
\subsection{Evaluation Metric}
Similar to AudioBench~\cite{wang-etal-2025-audiobench} and VoiceBench~\cite{chen2024voicebench}, we use a Model-as-Judge evaluation method. Prior work such as PEDANTS~\cite{li-etal-2024-pedants} shows that model-based evaluation, especially with strong GPT-level judges, aligns closely with human judgment when applied correctly. We therefore use GPT-4.1 as the judge model. Each evaluation sample is scored independently on a 0--5 scale, and we report the mean score. Judging prompts are task-specific and shared in the supplementary material. 

\vspace{-2pt}
\subsection{Experimental Setup}
All experiments were conducted on 4 NVIDIA H100 GPUs with an effective batch size of 288 (24 per GPU with gradient accumulation of 3). We train using DeepSpeed ZeRO Stage~2 and BF16 mixed precision. We use a linear warm-up for the first 400 steps followed by cosine annealing with a peak learning rate of $5.6\times10^{-5}$. Optimization is performed with AdamW ($\beta_1=0.9$, $\beta_2=0.999$) and early stopping with a patience of 6 validation checks (2000 steps). We initialize the Whisper encoder, projector, and LLM from pretrained HuggingFace checkpoints. To ensure fair comparison, all systems are evaluated using the same prompts, judge model, and scoring protocol.

\renewcommand{\arraystretch}{1.00}

\begin{table*}[t]
  \caption{GPT-4.1 Model-as-Judge scores for open-ended instruction following and close-ended spoken QA. Higher is better ($\uparrow$). To keep comparison fair, if a model does not support a language, the score has been left blank. EN-DiVA and ML-DiVA are matched DiVA~\cite{held-etal-2025-distilling} baseline models trained on English-only ASR data and Multilingual ASR data respectively from our pool of training data.}
  \vspace{-4pt}
  \label{tab:main_results_allinone}
  \centering
  \footnotesize
  \setlength{\tabcolsep}{2.0pt}
  \renewcommand{\arraystretch}{1.1}
  \begin{tabular}{L{4.36cm}|C{0.74cm}C{0.74cm}C{0.74cm}C{0.74cm}C{0.74cm}C{0.90cm}|C{0.90cm} |C{0.74cm}C{0.74cm}C{0.90cm}|C{0.74cm}C{0.74cm}C{0.90cm}}
    \toprule
    \multirow{3}{*}{\textbf{Model}} &
    \multicolumn{6}{c|}{\textbf{Close-ended (A-MLQA)} ($\uparrow$)} &
    \multicolumn{7}{c}{\textbf{Open-ended} ($\uparrow$)} \\
    \cmidrule(lr){2-7}\cmidrule(lr){8-14}
    & \multirow{2}{*}{\textbf{EN}} & \multirow{2}{*}{\textbf{VI}} & \multirow{2}{*}{\textbf{ES}} & \multirow{2}{*}{\textbf{DE}} & \multirow{2}{*}{\textbf{ZH}} & \multirow{2}{*}{\textbf{Avg}}
    & \textbf{ZH} 
    & \multicolumn{3}{c|}{\textbf{EN}}
    & \multicolumn{3}{c}{\textbf{ID}} \\
    & & & & & & & \textbf{AE} & \textbf{AE} & \textbf{OH} & \textbf{Avg} & \textbf{AE} & \textbf{OH} & \textbf{Avg}\\
    \midrule

    \addlinespace[-3pt]
    \multicolumn{1}{l|}{\scriptsize\textit{\textbf{Text-only reference}}} &
    \multicolumn{6}{c|}{} &
    \multicolumn{1}{c|}{} &
    \multicolumn{3}{c|}{} &
    \multicolumn{3}{c}{} \\
    \addlinespace[-2pt]
    SEA-LION-v3-8B-IT &
    4.34 & 4.10 & 4.19 & 4.08 & 3.98 & 4.14 &
    4.21 & 4.59 & 4.44 & 4.52 & 4.19 & 3.85 & 4.02 \\
    \midrule

    \addlinespace[-3pt]
    \multicolumn{1}{l|}{\scriptsize\textit{\textbf{Cascaded}}}  &
    \multicolumn{6}{c|}{} &
    \multicolumn{1}{c|}{} &
    \multicolumn{3}{c|}{} &
    \multicolumn{3}{c}{} \\
    \addlinespace[-2pt]
    Whisper-v3 + SEA-LION-v3-8B-IT &
    4.20 & 3.75 & 4.09 & 4.01 & 3.94 & 3.99 &
    3.95 & 3.99 & 4.17 & 4.08 & 3.68 & 3.61 & 3.65 \\
    \midrule

    \addlinespace[-3pt]
    \multicolumn{1}{l|}{\scriptsize\textit{\textbf{Zero-shot end-to-end}}} &
    \multicolumn{6}{c|}{} &
    \multicolumn{1}{c|}{} &
    \multicolumn{3}{c|}{} &
    \multicolumn{3}{c}{} \\
    \addlinespace[-2pt]
    Glm-4-voice~\cite{zeng2024glm} &
    3.04 & -- & -- & -- & 2.86 & 2.95 &
    3.41 & 3.62 & 3.72 & 3.67 & -- & -- & -- \\
    MERaLiON-2-10B~\cite{he2025meralion} &
    1.73 & 0.90 & 1.28 & 1.31 & 1.32 & 1.31 &
    1.96 & 4.59 & 4.06 & 4.33 & 3.59 & 3.06 & 3.33 \\
    SeaLLMs-Audio~\cite{SeaLLMs-Audio} &
    3.74 & 2.66 & 2.60 & 2.54 & 3.41 & 2.99 &
    3.72 & 3.93 & 3.37 & 3.65 & 3.78 & 3.27 & 3.53 \\
    Qwen2-Audio~\cite{chu2024qwen2} &
    3.54 & -- & 3.37 & 3.09 & 3.38 & 3.02 &
    3.65 & 3.77 & 3.65 & 3.71 & -- & -- & -- \\ 
    \midrule

    \addlinespace[-3pt]
    \multicolumn{1}{l|}{\scriptsize\textit{\textbf{Distilled end-to-end}}} &
    \multicolumn{6}{c|}{} &
    \multicolumn{1}{c|}{} &
    \multicolumn{3}{c|}{} &
    \multicolumn{3}{c}{} \\
    \addlinespace[-2pt]
    EN-DiVA~\cite{held-etal-2025-distilling} (re-trained baseline) &
    4.12 & -- & -- & -- & -- & 4.12 &
    -- & 3.57 & 3.28 & 3.43 & -- & -- & -- \\
    ML-DiVA~\cite{held-etal-2025-distilling} (re-trained baseline) &
    4.09 & 3.64 & 4.02 & 3.97 & 3.55 & 3.85 &
    2.87 & 4.28 & 4.03 & 4.16 & 2.96 & 3.11 & 3.04 \\
    \textbf{Ours (soft-gating)} &
    \textbf{4.12} & \textbf{3.68} & \textbf{4.03} & \textbf{4.02} & \textbf{3.56} & \textbf{3.88} \textcolor{teal}{\tiny$\uparrow$1} &
    \textbf{3.21} \textcolor{teal}{\tiny$\uparrow$12}& \textbf{4.58} & \textbf{4.29} & \textbf{4.44} \textcolor{teal}{\tiny$\uparrow$7} & \textbf{3.66} & \textbf{3.58} & \textbf{3.62} \textcolor{teal}{\tiny$\uparrow$19}\\
    \textbf{Ours (hard-gating)} &
    \textbf{4.16} & \textbf{3.72} & \textbf{4.13} & \textbf{4.07} & \textbf{3.70} & \textbf{3.96} \textcolor{teal}{\tiny$\uparrow$3}&
    \textbf{3.33} \textcolor{teal}{\tiny$\uparrow$16} & \textbf{4.39} & \textbf{4.42} & \textbf{4.41} \textcolor{teal}{\tiny$\uparrow$6} & \textbf{3.76} & \textbf{3.66} & \textbf{3.71} \textcolor{teal}{\tiny$\uparrow$22}\\
    \bottomrule
  \end{tabular}
  \vspace{4pt}
  \\ 
  {Languages: \textit{EN} = English, \textit{ID} = Indonesian, \textit{ZH} = Chinese-Simplified, \textit{VI} = Vietnamese, \textit{ES} = Spanish, \textit{DE} = German \\
  \footnotesize{Datasets: \textit{AE} = Audio-AlpacaEval, \textit{OH} = Audio-OpenHermes, \textit{A-MLQA} = Audio-MLQA \\
  }}
  \vspace{-12pt}
\end{table*}

\vspace{-2pt}
\section{Results}


In the cascaded baseline system, the LLM directly receives the transcription of the speech instruction from whisper-large-v3. We also establish the two matched baselines EN-DiVA (\emph{which is closest to the original DiVA model}) and ML-DiVA trained on only the English subset and the entire training dataset respectively. Simply scaling data and adding additional languages (notably similar languages such as German and Spanish) improves open-ended EN performance by 21\% (ML-DiVA), likely due to (i) shared linguistic structure and (ii) increased data diversity and volume. Next we detail gains obtained using our approach.

\subsection{Open-Ended Evaluation}
For this task, all models are evaluated without a preferred reference answer, and the MaJ (GPT-4.1) evaluates the subjective correctness of the response. As expected, the text-only LLM (SEA-LION-v3-8B-IT) acts as the performance upper bound of the proposed method across all languages. Our best approach (hard-gating) achieves an average gain of 14\% over ML-DiVA on instruction following. Notably, in the Indonesian (ID) task, our model improves the average score from $3.04$ (ML-DiVA) to $3.71$, demonstrating the effectiveness of language-aware routing in protecting lower-represented languages from interference. Our approach also performs better than existing Speech LLMs for all languages other than ZH (least similar language).

\subsection{Close-Ended Evaluation}
On Audio-MLQA, our model improves over strong baselines~\cite{SeaLLMs-Audio} and~\cite{Qwen-Audio} by 32\% and 31\%. This result is expected for close-ended QA, where performance relies heavily on speech--text alignment. However, language-aware distillation still provides consistent additional gains, as the hard-gating variant reaches a close-ended average of $3.96$, a $\sim$3\% improvement over ML-DiVA and close to the text-only reference ($4.14$). In contrast, some SFT-based models (e.g., MERaLiON-2-10B) sometimes fail to align the audio question with the textual context and default to generic “answer not found” responses, indicating weaker generalization to varied task instructions.

\vspace{-2pt}
\subsection{Ablation Studies}
\label{sec:ablation_studies}
\begin{table}[t]
  \caption{Ablations on query length $L$ and gating design. We report validation losses for input and output distillation ($\mathcal{L}_{\text{IN}}$, $\mathcal{L}_{\text{OUT}}$); for gating variants we additionally report LID accuracy.}
  \vspace{-4pt}
  \label{tab:ablation_gating}
  \centering
  \footnotesize
  \setlength{\tabcolsep}{4pt}
  \renewcommand{\arraystretch}{1.0}
  \begin{tabular}{lccc}
    \toprule
    \textbf{Variant} & $\boldsymbol{\mathcal{L}_{\text{OUT}}}$ $\downarrow$ & $\boldsymbol{\mathcal{L}_{\text{IN}}}$ $\downarrow$ & LID Acc. $\uparrow$ \\
    \midrule
    $L=64$  & 57.92 & 8.63 & -- \\
    $L=128$ & 40.69 & 1.16 & -- \\
    {$\mathbf{L}=\mathbf{256}$} & \textbf{39.47} & \textbf{0.97} & -- \\
    \midrule
    \hspace{4pt}\textbf{\small + Conv. DS + Pool Gate} & 37.31 & \textbf{0.84} & \textbf{95.15\%} \\
    \hspace{4pt}{\small + Attn. Pool Gate} & \textbf{36.93} & 0.94 & 94.97\% \\
    \bottomrule
  \end{tabular}
  \vspace{-14pt}
\end{table}

We perform ablation studies to evaluate the impact of query length $L$, routing modes, and gating network architectures on model performance (Table~\ref{tab:ablation_gating}). First, we evaluate query length $L$ using a static shared sequence; Increasing $L$ from 64 to 256 reduces input distillation loss by 89\% ($8.63 \rightarrow 0.97$), suggesting that higher capacity is important for capturing complex phonetic--semantic mappings. Second, both proposed language-aware gating mechanisms (Conv. DS and Attn. Pool) show strong performance, achieving language identification (LID) accuracy exceeding 94.9\%. This supports accurate routing of speech features and indicates that dynamic query selection facilitates cross-modal alignment regardless of the specific gating architecture. Finally, we find that hard query selection consistently outperforms soft mixing in downstream tasks (shown in Table~\ref{tab:main_results_allinone}). This suggests that hard-gating provides stronger decoupling of language-specific information and avoids an ``averaging'' effect where dominant languages can still interfere with lower-represented ones during retrieval.


\vspace{-4pt}
\section{Conclusion}

In this study, we introduce a language-aware distillation framework to resolve the language interference bottleneck in distilled multilingual Speech LLMs. By replacing static query sequences with a dynamic query bank and gating mechanism, we effectively disentangle language-specific information while maintaining high data efficiency. Our results demonstrate that this architecture allows performant multilingual interaction using only 5.8K hours of ASR data while keeping the backbone speech encoder and LLM frozen. This approach offers a scalable and resource-efficient paradigm for extending advanced speech understanding to a broader range of global languages.

\ifcameraready
\section{Acknowledgments}
This research was supported by the WeBank-NTU Joint Research Institute on FinTech, Nanyang Technological University, Singapore. The computational work for this article was partially performed on resources of the National Supercomputing Centre, Singapore (https://www.nscc.sg) and partially supported by the High Performance Computing Centre of Nanyang Technological University, Singapore.
\else
\fi

\section{Generative AI Use Disclosure}
Generative AI tools were used for limited assistance with manuscript editing and presentation (e.g., grammatical validation, removal of redundant sentences and phrases, preparing LaTeX equations and LaTeX formatting suggestions). The literature review, and all scientific contributions including but not limited to problem formulation, methodology, experiments, results, and conclusions were performed by the authors. All authors reviewed the content and are responsible for the final submission.

\bibliographystyle{IEEEtran}
\bibliography{mybib}

@inproceedings{capleaf_vivoice_2024,
  author    = {Capleaf},
  title = {{viVoice}: Enabling Vietnamese Multi-Speaker Speech Synthesis},
  year      = {2024},
  booktitle = {Hugging Face},
  url       = {https://huggingface.co/datasets/capleaf/viVoice},
  urldate   = {2026-02-21},
  note      = {License: CC BY-NC-SA 4.0 (gated access)}
}

@inproceedings{li2023yodas,
  title={Yodas: Youtube-Oriented Dataset for Audio and Speech},
  author={Li, Xinjian and Takamichi, Shinnosuke and Saeki, Takaaki and Chen, William and Shiota, Sayaka and Watanabe, Shinji},
  booktitle={2023 IEEE Automatic Speech Recognition and Understanding Workshop (ASRU)},
  pages={1--8},
  year={2023},
  organization={IEEE}
}

@inproceedings{held-etal-2025-distilling,
    title = "Distilling an End-to-End Voice Assistant Without Instruction Training Data",
    author = "Held, William  and
      Zhang, Yanzhe  and
      Li, Minzhi  and
      Shi, Weiyan  and
      Ryan, Michael J  and
      Yang, Diyi",
    editor = "Che, Wanxiang  and
      Nabende, Joyce  and
      Shutova, Ekaterina  and
      Pilehvar, Mohammad Taher",
    booktitle = "Proceedings of the 63rd Annual Meeting of the Association for Computational Linguistics (Volume 1: Long Papers)",
    month = jul,
    year = "2025",
    address = "Vienna, Austria",
    publisher = "Association for Computational Linguistics",
    url = "https://aclanthology.org/2025.acl-long.388/",
    doi = "10.18653/v1/2025.acl-long.388",
    pages = "7876--7891",
    ISBN = "979-8-89176-251-0",
}

@article{chu2024qwen2,
  title={Qwen2-audio Technical Report},
  author={Chu, Yunfei and Xu, Jin and others},
  journal={arXiv preprint arXiv:2407.10759},
  year={2024}
}

@article{Qwen-Audio,
  title={Qwen-Audio: Advancing Universal Audio Understanding via Unified Large-Scale Audio-Language Models},
  author={Chu, Yunfei and Xu, Jin and others},
  journal={arXiv preprint arXiv:2311.07919},
  year={2023}
}

@misc{SeaLLMs-Audio,
      title={SeaLLMs-Audio: Large Audio-Language Models for Southeast Asia}, 
      author={Chaoqun Liu and Mahani Aljunied and others},
      year={2025},
      eprint={2511.01670},
      archivePrefix={arXiv},
      url={https://arxiv.org/abs/2511.01670}, 
}

@inproceedings{he2025meralion,
    title = "{MER}a{L}i{ON}-{A}udio{LLM}: Advancing Speech and Language Understanding for {S}ingapore",
    author = "He, Yingxu  and
      Liu, Zhuohan  and
      Lin, Geyu  and
      Sun, Shuo  and
      Wang, Bin  and
      Zhang, Wenyu  and
      Zou, Xunlong  and
      Chen, Nancy F.  and
      Aw, AiTi",
    editor = "Mishra, Pushkar  and
      Muresan, Smaranda  and
      Yu, Tao",
    booktitle = "Proceedings of the 63rd Annual Meeting of the Association for Computational Linguistics (Volume 3: System Demonstrations)",
    month = jul,
    year = "2025",
    address = "Vienna, Austria",
    publisher = "Association for Computational Linguistics",
    url = "https://aclanthology.org/2025.acl-demo.3/",
    doi = "10.18653/v1/2025.acl-demo.3",
    pages = "22--30",
    ISBN = "979-8-89176-253-4",
}

@inproceedings{
    tang2024salmonn,
    title={{SALMONN}: Towards Generic Hearing Abilities for Large Language Models},
    author={Changli Tang and Wenyi Yu and Guangzhi Sun and Xianzhao Chen and Tian Tan and Wei Li and Lu Lu and Zejun MA and Chao Zhang},
    booktitle={The Twelfth International Conference on Learning Representations},
    year={2024},
    url={https://openreview.net/forum?id=14rn7HpKVk}
}

@inproceedings{fang-etal-2024-llama-omni,
  title={LLaMA-omni 2: LLM-based real-time spoken chatbot with autoregressive streaming speech synthesis},
  author={Fang, Qingkai and Zhou, Yan and Guo, Shoutao and Zhang, Shaolei and Feng, Yang},
  booktitle={Proceedings of the 63rd Annual Meeting of the Association for Computational Linguistics (Volume 1: Long Papers)},
  pages={18617--18629},
  year={2025}
}

@article{grattafiori2024llama,
  title={The llama 3 herd of models},
  author={Grattafiori, Aaron and Dubey, Abhimanyu and Jauhri, Abhinav and Pandey, Abhinav and Kadian, Abhishek and Al-Dahle, Ahmad and Letman, Aiesha and Mathur, Akhil and Schelten, Alan and Vaughan, Alex and others},
  journal={arXiv preprint arXiv:2407.21783},
  year={2024}
}

@inproceedings{ng2025sea,
    title = "{SEA}-{LION}: {S}outheast {A}sian Languages in One Network",
    author = "Ng, Raymond and Nguyen, Thanh Ngan and others",
    booktitle = "Proceedings of the 14th International Joint Conference on Natural Language Processing and the 4th Conference of the Asia-Pacific Chapter of the Association for Computational Linguistics",
    month = dec,
    year = "2025",
    address = "Mumbai, India",
    publisher = "The Asian Federation of Natural Language Processing and The Association for Computational Linguistics",
    url = "https://aclanthology.org/2025.ijcnlp-long.30/",
    doi = "10.18653/v1/2025.ijcnlp-long.30",
    pages = "512--526",
    ISBN = "979-8-89176-298-5",
}

@article{wang2023blsp,
  title={Blsp: Bootstrapping language-speech pre-training via behavior alignment of continuation writing},
  author={Wang, Chen and Liao, Minpeng and Huang, Zhongqiang and Lu, Jinliang and Wu, Junhong and Liu, Yuchen and Zong, Chengqing and Zhang, Jiajun},
  journal={arXiv preprint arXiv:2309.00916},
  year={2023}
}

@inproceedings{li-etal-2023-adaptive-gating,
    title = "Adaptive Gating in Mixture-of-Experts based Language Models",
    author = "Li, Jiamin  and
      Su, Qiang  and
      Yang, Yitao  and
      Jiang, Yimin  and
      Wang, Cong  and
      Xu, Hong",
    editor = "Bouamor, Houda  and
      Pino, Juan  and
      Bali, Kalika",
    booktitle = "Proceedings of the 2023 Conference on Empirical Methods in Natural Language Processing",
    month = dec,
    year = "2023",
    address = "Singapore",
    publisher = "Association for Computational Linguistics",
    url = "https://aclanthology.org/2023.emnlp-main.217/",
    doi = "10.18653/v1/2023.emnlp-main.217",
    pages = "3577--3587",
}

@inproceedings{wang2023language,
  title     = {{Language-Routing Mixture of Experts for Multilingual and Code-Switching Speech Recognition}},
  author    = {Wenxuan Wang and Guodong Ma and Yuke Li and Binbin Du},
  year      = {2023},
  booktitle = {{Interspeech 2023}},
  pages     = {1389--1393},
  doi       = {10.21437/Interspeech.2023-2292},
  issn      = {2958-1796},
}

@inproceedings{wang-etal-2025-audiobench,
    title = "{A}udio{B}ench: A Universal Benchmark for Audio Large Language Models",
    author = "Wang, Bin  and
      Zou, Xunlong  and others",
    booktitle = "Proceedings of the 2025 Conference of the Nations of the Americas Chapter of the Association for Computational Linguistics: Human Language Technologies (Volume 1: Long Papers)",
    month = apr,
    year = "2025",
    address = "Albuquerque, New Mexico",
    publisher = "Association for Computational Linguistics",
    url = "https://aclanthology.org/2025.naacl-long.218/",
    doi = "10.18653/v1/2025.naacl-long.218",
    pages = "4297--4316",
    ISBN = "979-8-89176-189-6",
}

@article{chen2024voicebench,
    title = "{V}oice{B}ench: Benchmarking {LLM}-Based Voice Assistants",
    author = "Chen, Yiming  and
      Yue, Xianghu  and others",
    journal = "Transactions of the Association for Computational Linguistics",
    volume = "14",
    year = "2026",
    address = "Cambridge, MA",
    publisher = "MIT Press",
    url = "https://aclanthology.org/2026.tacl-1.18/",
    doi = "10.1162/tacl.a.628",
    pages = "378--398",
}

@inproceedings{hsiao2025forgetting,
  title     = {{Analyzing Mitigation Strategies for Catastrophic Forgetting in End-to-End Training of Spoken Language Models}},
  author    = {Chi-Yuan Hsiao and Ke-Han Lu and Kai-Wei Chang and Chih-Kai Yang and Wei-Chih Chen and Hung-yi Lee},
  year      = {2025},
  booktitle = {{Interspeech 2025}},
  pages     = {3234--3238},
  doi       = {10.21437/Interspeech.2025-409},
  issn      = {2958-1796},
}

@inproceedings{
majumder2024tango,
title={Tango 2: Aligning Diffusion-based Text-to-Audio Generative Models through Direct Preference Optimization},
author={Navonil Majumder and Chia-Yu Hung and Deepanway Ghosal and Wei-Ning Hsu and Rada Mihalcea and Soujanya Poria},
booktitle={ACM Multimedia 2024},
year={2024},
url={https://openreview.net/forum?id=7lqptq5dLG}
}

@inproceedings{noroozi2024instruction,
  title     = {{Instruction Data Generation and Unsupervised Adaptation for Speech Language Models}},
  author    = {Vahid Noroozi and Zhehuai Chen and Somshubra Majumdar and Steve Huang and Jagadeesh Balam and Boris Ginsburg},
  year      = {2024},
  booktitle = {{Interspeech 2024}},
  pages     = {4049--4053},
  doi       = {10.21437/Interspeech.2024-1575},
  issn      = {2958-1796},
}

@inproceedings{dao2025speechless,
  title     = {{Speechless: Speech Instruction Training Without Speech for Low Resource Languages}},
  author    = {Alan Dao and Dinh Bach Vu and Huy Hoang Ha and Tuan Le Duc Anh and Shreyas Gopal and Yue Heng Yeo and Warren Keng Hoong Low and Eng Siong Chng and Jia Qi Yip},
  year      = {2025},
  booktitle = {{Interspeech 2025}},
  pages     = {3239--3243},
  doi       = {10.21437/Interspeech.2025-1292},
  issn      = {2958-1796},
}

@INPROCEEDINGS{chen2024salm,
  author={Chen, Zhehuai and Huang, He and Andrusenko, Andrei and Hrinchuk, Oleksii and Puvvada, Krishna C. and Li, Jason and Ghosh, Subhankar and Balam, Jagadeesh and Ginsburg, Boris},
  booktitle={ICASSP 2024 - 2024 IEEE International Conference on Acoustics, Speech and Signal Processing (ICASSP)}, 
  title={SALM: Speech-Augmented Language Model with in-Context Learning for Speech Recognition and Translation}, 
  year={2024},
  volume={},
  number={},
  pages={13521-13525},
  doi={10.1109/ICASSP48485.2024.10447553}}

@InProceedings{whisper,
  title = 	 {Robust Speech Recognition via Large-Scale Weak Supervision},
  author =       {Radford, Alec and Kim, Jong Wook and Xu, Tao and Brockman, Greg and Mcleavey, Christine and Sutskever, Ilya},
  booktitle = 	 {Proceedings of the 40th International Conference on Machine Learning},
  pages = 	 {28492--28518},
  year = 	 {2023},
  editor = 	 {Krause, Andreas and Brunskill, Emma and Cho, Kyunghyun and Engelhardt, Barbara and Sabato, Sivan and Scarlett, Jonathan},
  volume = 	 {202},
  series = 	 {Proceedings of Machine Learning Research},
  month = 	 {23--29 Jul},
  publisher =    {PMLR},
  url = 	 {https://proceedings.mlr.press/v202/radford23a.html},
}

@article{zeng2024glm,
  title={Glm-4-voice: Towards intelligent and human-like end-to-end spoken chatbot},
  author={Zeng, Aohan and Du, Zhengxiao and Liu, Mingdao and Wang, Kedong and Jiang, Shengmin and Zhao, Lei and Dong, Yuxiao and Tang, Jie},
  journal={arXiv preprint arXiv:2412.02612},
  year={2024}
}

@ARTICLE{aaron_2stage,
  author={Kwok, Chin Yuen and Liu, Hexin and Yip, Jia Qi and Li, Sheng and Chng, Eng Siong},
  journal={IEEE Transactions on Audio, Speech and Language Processing}, 
  title={A Two-Stage LoRA Strategy for Expanding Language Capabilities in Multilingual ASR Models}, 
  year={2025},
  volume={33},
  number={},
  pages={2576-2590},
  doi={10.1109/TASLPRO.2025.3578752}}

@inproceedings{li2023blip,
author = {Li, Junnan and Li, Dongxu and Savarese, Silvio and Hoi, Steven},
title = {BLIP-2: bootstrapping language-image pre-training with frozen image encoders and large language models},
year = {2023},
publisher = {JMLR.org},
booktitle = {Proceedings of the 40th International Conference on Machine Learning},
articleno = {814},
numpages = {13},
location = {Honolulu, Hawaii, USA},
series = {ICML'23}
}

@inproceedings{dong2024_interspeech,
  title     = {{Integrating Speech Self-Supervised Learning Models and Large Language Models for ASR}},
  author    = {Ling Dong and Zhengtao Yu and Wenjun Wang and Yuxin Huang and Shengxiang Gao and Guojiang Zhou},
  year      = {2024},
  booktitle = {{Interspeech 2024}},
  pages     = {3954--3958},
  doi       = {10.21437/Interspeech.2024-1760},
  issn      = {2958-1796},
}

@inproceedings{shang2024_interspeech,
  title     = {{An End-to-End Speech Summarization Using Large Language Model}},
  author    = {Hengchao Shang and Zongyao Li and Jiaxin Guo and Shaojun Li and Zhiqiang Rao and Yuanchang Luo and Daimeng Wei and Hao Yang},
  year      = {2024},
  booktitle = {{Interspeech 2024}},
  pages     = {1950--1954},
  doi       = {10.21437/Interspeech.2024-1428},
  issn      = {2958-1796},
}

@inproceedings{li-etal-2024-pedants,
    title = "{PEDANTS}: Cheap but Effective and Interpretable Answer Equivalence",
    author = "Li, Zongxia  and
      Mondal, Ishani  and
      Nghiem, Huy  and
      Liang, Yijun  and
      Boyd-Graber, Jordan",
    editor = "Al-Onaizan, Yaser  and
      Bansal, Mohit  and
      Chen, Yun-Nung",
    booktitle = "Findings of the Association for Computational Linguistics: EMNLP 2024",
    month = nov,
    year = "2024",
    address = "Miami, Florida, USA",
    publisher = "Association for Computational Linguistics",
    url = "https://aclanthology.org/2024.findings-emnlp.548/",
    doi = "10.18653/v1/2024.findings-emnlp.548",
    pages = "9373--9398",
}

@inproceedings{lewis2019mlqa,
  title={MLQA: Evaluating cross-lingual extractive question answering},
  author={Lewis, Patrick and Oguz, Barlas and Rinott, Ruty and Riedel, Sebastian and Schwenk, Holger},
  booktitle={Proceedings of the 58th annual meeting of the association for computational linguistics},
  pages={7315--7330},
  year={2020}
}

@inproceedings{yan2025uro,
    title = "{URO}-Bench: Towards Comprehensive Evaluation for End-to-End Spoken Dialogue Models",
    author = "Yan, Ruiqi  and
      Li, Xiquan and others",
    booktitle = "Findings of the Association for Computational Linguistics: EMNLP 2025",
    month = nov,
    year = "2025",
    address = "Suzhou, China",
    publisher = "Association for Computational Linguistics",
    url = "https://aclanthology.org/2025.findings-emnlp.933/",
    doi = "10.18653/v1/2025.findings-emnlp.933",
    pages = "17211--17242",
    ISBN = "979-8-89176-335-7",
}

@misc{magicdata_2019,
  author       = {{Magic Data Technology Co., Ltd.}},
  title        = {Magic Data Open Source Corpus (ID: 101)},
  howpublished = {\url{http://www.imagicdatatech.com/index.php/home/dataopensource/data_info/id/101}},
  month        = may,
  year         = {2019},
  note         = {Accessed: 2026-02-25}
}

@inproceedings{commonvoice:2020,
    title = "Common Voice: A Massively-Multilingual Speech Corpus",
    author = "Ardila, Rosana  and
      Branson, Megan  and others",
    booktitle = "Proceedings of the Twelfth Language Resources and Evaluation Conference",
    month = may,
    year = "2020",
    address = "Marseille, France",
    publisher = "European Language Resources Association",
    url = "https://aclanthology.org/2020.lrec-1.520/",
    pages = "4218--4222",
    language = "eng",
    ISBN = "979-10-95546-34-4",
}

@inproceedings{wang2024blsp_emo,
  title={Blsp-emo: Towards empathetic large speech-language models},
  author={Wang, Chen and Liao, Minpeng and others},
  booktitle={Proceedings of the 2024 Conference on Empirical Methods in Natural Language Processing},
  pages={19186--19199},
  year={2024}
}

@inproceedings{wang2025benchmarking,
    title = "Benchmarking Contextual and Paralinguistic Reasoning in Speech-{LLM}s: A Case Study with In-the-Wild Data",
    author = "Wang, Qiongqiong  and
      Sailor, Hardik Bhupendra  and others",
    booktitle = "Findings of the Association for Computational Linguistics: EMNLP 2025",
    month = nov,
    year = "2025",
    address = "Suzhou, China",
    publisher = "Association for Computational Linguistics",
    url = "https://aclanthology.org/2025.findings-emnlp.760/",
    doi = "10.18653/v1/2025.findings-emnlp.760",
    pages = "14133--14148",
    ISBN = "979-8-89176-335-7",
}

@article{nguyen2023generative,
  title={Generative spoken dialogue language modeling},
  author={Nguyen, Tu Anh and Kharitonov, Eugene and Copet, Jade and Adi, Yossi and Hsu, Wei-Ning and Elkahky, Ali and Tomasello, Paden and Algayres, Robin and Sagot, Benoit and Mohamed, Abdelrahman and others},
  journal={Transactions of the Association for Computational Linguistics},
  volume={11},
  pages={250--266},
  year={2023},
  publisher={MIT Press One Broadway, 12th Floor, Cambridge, Massachusetts 02142, USA~…}
}

\end{document}